\pdfoutput=1
\documentclass[11pt]{article}

\usepackage[]{acl}

\usepackage{times}
\usepackage{latexsym}
\usepackage{booktabs}       % professional-quality tables
\usepackage{amsfonts}       % blackboard math symbols
\usepackage{nicefrac}       % compact symbols for 1/2, etc.
\usepackage{microtype}      % microtypography
\usepackage{xcolor}         % colors
\usepackage{geometry}                 %更改页边距所需
\usepackage{booktabs}                 %表格线条加粗所需
\usepackage{multirow}				  %合并多行
\usepackage{graphicx}
\usepackage{enumitem}
\usepackage{enumerate}
\usepackage{wrapfig}
\usepackage{multicol}

\usepackage{pifont}% http://ctan.org/pkg/pifont

\usepackage[T1]{fontenc}
\usepackage[utf8]{inputenc}

\usepackage{microtype}

\title{KMIR: A Benchmark for Evaluating Knowledge Memorization, Identification and Reasoning Abilities of Language Models}

\author{Daniel Gao\textsuperscript{1}, Yantao Jia\textsuperscript{1}, Lei Li\textsuperscript{1}, Chengzhen Fu\textsuperscript{1}, Zhicheng Dou\textsuperscript{2},\\ \textbf{Hao Jiang\textsuperscript{1}, Xinyu Zhang\textsuperscript{1}, Lei Chen\textsuperscript{3}, Zhao Cao\textsuperscript{1}}\\
\textsuperscript{1}Huawei Poisson Laboratory, Beijing, China\\
\textsuperscript{2}Gaoling School of Artificial Intelligence, Renmin University of China, Beijing, China\\
\textsuperscript{3}Hong Kong University of Science and Technology, Hong Kong, China\\
{\tt {\small dangao1900@163.com, \textbraceleft jiayantao, fuchengzhen, lilei229, jianghao66, zhangxinyu35, caozhao1\textbraceright@huawei.com,}} \\ {\tt {\small dou@ruc.edu.cn, leichen@cse.ust.hk
}}

}

\begin{document}
\maketitle

\begin{abstract}
Previous works show the great potential of pre-trained language models (PLMs) for storing a large amount of factual knowledge. However, to figure out whether PLMs can be reliable knowledge sources and used as alternative knowledge bases (KBs), we need to further explore some critical features of PLMs. Firstly, \textbf{knowledge memorization and identification abilities}: traditional KBs can store various types of entities and relationships; do PLMs have a high knowledge capacity to store different types of knowledge? Secondly, \textbf{reasoning ability}: a qualified knowledge source should not only provide a collection of facts, but support a symbolic reasoner. Can PLMs derive new knowledge based on the correlations between facts? To evaluate these features of PLMs, we propose a benchmark, named \textbf{K}nowledge \textbf{M}emorization, \textbf{I}dentification, and \textbf{R}easoning test (KMIR). KMIR covers 3 types of knowledge, including general knowledge, domain-specific knowledge, and commonsense, and provides 184,348 well-designed questions. Preliminary experiments with various representative pre-training language models on KMIR reveal many interesting phenomenons:  1) The memorization ability of PLMs depends more on the number of parameters than training schemes. 2) Current PLMs are struggling to robustly remember the facts. 3) Model compression technology retains the amount of knowledge well, but hurts the identification and reasoning abilities. We hope KMIR can facilitate the design of PLMs as better knowledge sources.

\end{abstract}

\section{Introduction}

Language models pre-trained on large-scale unstructured documents have proven extremely powerful on many Natural Language Processing (NLP) tasks~\cite{brown2020language, reimers2019sentence, raffel2019exploring, sanh2019distilbert, peters2018deep, dai2019transformer}. Many recent inspiring works~\cite{Petroni2019LanguageMA, Petroni2021KILTAB} further reveal the potential of the pre-trained language models (PLM) to be an alternative to structured knowledge bases~\cite{lenat1995cyc, Bollacker2008FreebaseAC, Auer2007DBpediaAN, Vrandecic2014WikidataAF, Speer2017ConceptNet5A, mitchell2018never} in some cases, especially at avoiding large-scale schema engineering. However, if a PLM can be a reliable knowledge source, it must equip the essential features motivated from human cognition~\cite{hayes2014memory} as: (1) \textbf{Strong knowledge memorization and identification}: A PLM should precisely memorize and identify the knowledge like those stored in a KB to facilitate the downstream language tasks such as QA, dialogue. (2) \textbf{Powerful reasoning ability}: The knowledge in PLM should be able to investigate the correlations between knowledge and easily be invoked to infer new knowledge from existing ones.

\begin{figure}[t]
    % \vspace{-0.1cm}
    % \centering
    \includegraphics[width=7.6cm]{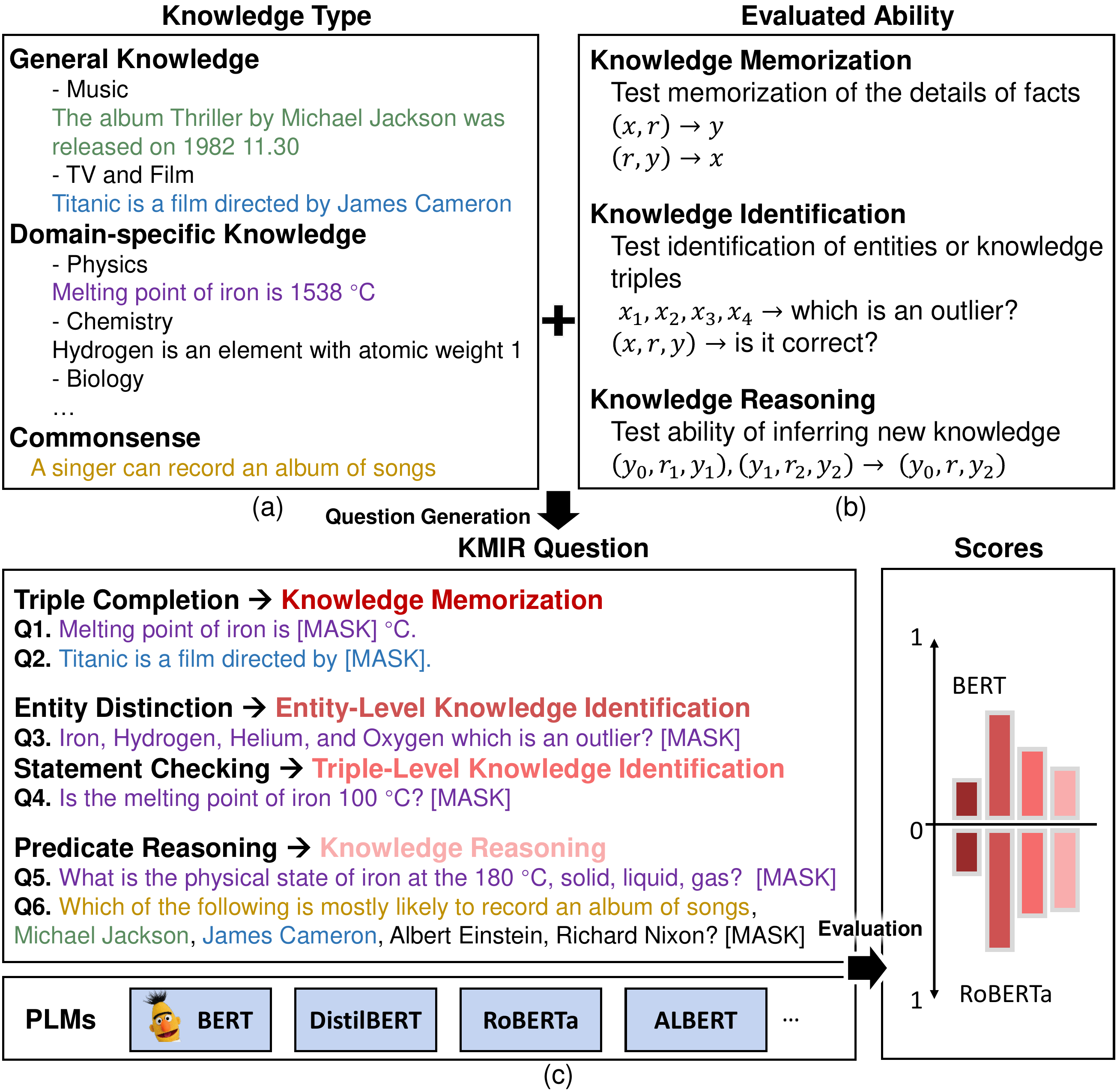}
    % \vspace{-0.3cm}
    \caption{KMIR covers multiple types of knowledge with diverse types of questions to comprehensively evaluate the knowledge memorization, identification and reasoning abilities of pre-trained language models.}
    \label{kct_domains}
    % \vspace{-0.3cm}
\end{figure}

Previous benchmarks, such as GLUE~\cite{wang2018glue} and SuperGLUE~\cite{wang2019superglue}, mainly focus on evaluating the general language understanding abilities, rather than knowledge related abilities. Another pioneering work, LAMA~\cite{Petroni2019LanguageMA}, focuses on probing knowledge in language models. LAMA mainly evaluates the knowledge memorization by asking models questions in the form of triple completion, like \textbf{Q1} and \textbf{Q2} in Figure~\ref{kct_domains}. However, a qualified model is highly supposed to act more than memorizing knowledge. Can PLMs identify different entities (e.g.,  \textbf{Q3}), identify the correctness of a fact (e.g., \textbf{Q4}), or collect related triples to derive new knowledge (e.g., \textbf{Q5})?  Unfortunately, if we only examine by means of triple completion questions, these fundamental knowledge abilities for downstream tasks are hard to be evaluated.

Therefore, we propose a benchmark for \textbf{K}nowledge \textbf{M}emorization, \textbf{I}dentification, and \textbf{R}easoning test (KMIR), with the idea shown in Figure~\ref{kct_domains}. It is designed motivated by the examinations for humans, like SAT and GRE, which test different abilities with various subjects of knowledge. Specifically, KMIR covers diverse types of knowledge referred in ~\cite{davis2015commonsense}, and shown in Figure~\ref{kct_domains} (a), including general knowledge often referred to in human communication, domain-specific knowledge about elementary-level natural and social science, and commonsense of universally accepted beliefs about the world and obvious to most humans. And we collect 192,078 knowledge triples about these types of knowledge from publicly available knowledge sources, i.e, WikiData and ConceptNet.  

Besides, we propose a syllabus to summarize the evaluated knowledge-related abilities in three-fold: knowledge memorization, identification, and reasoning, as shown in Figure~\ref{kct_domains} (b). 
Then, according to the characteristics of these three abilities, a template-based question generation method is developed to automatically convert the knowledge triples into question-answer pairs as
four types of prompt-based cloze questions, including triplet completion, entity distinction, statement checking, and predicate reasoning, as shown in Figure~\ref{kct_domains} (c).  Finally, we construct a large-scale dataset with 184,348 questions\footnote{KMIR is avaliable at ~\href{github.com/KMIR2021/KMIR}{github.com/KMIR2021/KMIR}.}.

Finally, we evaluate several representative PLMs on the proposed questions, including BERT~\cite{lan2019albert}, ALBERT~\cite{lan2019albert}, RoBERTa~\cite{liu2019roberta}, etc. Some interesting phenomena are found: 1) The memorization abilities of PLMs depend more on the number of parameters than training schemes. 2) Model compression technology retains the amount of knowledge well, but hurts the reasoning ability. 3) Current PLMs are struggling to robustly remember the facts. The predictions of questions derived from same facts usually lack consistency. 4) We find PLMs can solve reasoning questions better in the case that the entities contained in the question are highly co-occurred with the correct answers in the documents of Wikipedia. This indicates models who can answer reasoning questions in that they just memorized context information by a large-amount of parameters.

In summary, our paper mainly has three contributions:

\begin{itemize}[leftmargin=*]
\setlength{\itemsep}{2pt}
\setlength{\parsep}{0pt}
\setlength{\parskip}{0pt}
\item We propose a systematic evaluation syllabus. It comprehensively measures the knowledge-related abilities of PLMs, including knowledge memorization, identification, and reasoning. Moreover, it covers diverse types of knowledge including general knowledge, commonsense as well as domain-specific knowledge.
% \dou{introduce this earlier}{}.

\item We develop a template-based method that can convert knowledge triples to diverse cloze questions, and obtain a large-scale dataset containing 184,348 QA samples. An intensive validation for quality control of samples is carried with more than 532 hours.

\item We evaluate several representative PLMs on KMIR, and find many interesting phenomena that could be useful for designing better PLMs.

\end{itemize}

\section{Related Work}

\textbf{Language Understanding Benchmarks.}
Previous NLP benchmarks are usually for evaluate general language understanding, such as slot filling~\cite{elsahar2019t, levy2017zero}, QA~\cite{yang2018hotpotqa, rajpurkar2016squad, joshi2017triviaqa, fan2019eli5, kwiatkowski2019natural, liu2020logiqa, ding2019cognitive, clark2019boolq, kassner2020pretrained}, dialogue~\cite{dinan2018wizard}, entailment~\cite{williams2018broad, rocktaschel2015reasoning, dagan2005pascal, morgenstern2015winograd}. 
%Slot filling tasks aim to measure the ability to collect information on certain relations of entities from a large scale of texts. 
For example, some question answering tasks aim to evaluate machine reading comprehension or reason over a knowledge source, such as Wikipedia. Some other QA benchmarks, e.g., LogiQA~\cite{liu2020logiqa} and HotpotQA~\cite{yang2018hotpotqa}, also evaluate the ability of deductive reasoning. These benchmarks aim to evaluate the abilities of an NLP model on a specific task. Recently, GLUE~\cite{wang2018glue} is proposed to measure the models' universal language understanding. It provides a collection of tasks set to comprehensively evaluate a model. Further, a more difficult version, SuperGLUE~\cite{wang2019superglue} is proposed to cope with the increasing performance of PLMs. LReasoner~\cite{wang2021logic} is another universal language understanding benchmark, but focuses more on the robustness of PLMs. 
Compared to the above benchmarks evaluating the PLMs in terms of general downstream language tasks, KMIR examines PLMs in terms of knowledge-related abilities with diverse types of knowledge, and somehow measures the qualities of PLMs as knowledge sources.  
% KMIR   analyze the quality of the model from the perspective of knowledge utilization
% KMIR is a diagnostic dataset which focuses on the knowledge related part of a PLM model rather than the whole pipeline of a specific task.

\textbf{Knowledge Probe Benchmark.} %  Pre-training Language Model
% \dou{should we put this sentence in 3.1?}{}
Recently, pre-trained language model~\cite{devlin2019bert, radford2019language, lewis2020bart} has made significant progress in the field of natural language processing and has gradually become one of the mainstream paradigms. 
%By pre-training the language model on the large unsupervised text corpus and then fine-tuning the specific downstream tasks, the pre-training language model achieves quite good results on all kinds of NLP tasks. 
A part of the research~\cite{Roberts2020HowMK, Petroni2019LanguageMA, Petroni2021KILTAB, jiang2020can, roberts2020much, yu2020jaket} has begun to explore the factual knowledge contained in pre-trained language models. LAMA~\cite{Petroni2019LanguageMA} is the first work to explore whether a language model learns and stores some factual knowledge by pre-training on a large text corpus. To address this problem, LAMA uses the triple completion task to probe whether a language model contains a certain knowledge. If a PLM can successfully predict the words that have been masked in a knowledge triple, then the PLM is assumed to include this knowledge. KILT~\cite{Petroni2021KILTAB} is another benchmark by testing the performance of PLMs on the downstream knowledge-intensive tasks, such as open-domain question answering, fact-checking, and slot filling.
% which evaluates the performance of a PLM on various types of tasks that require utilizing large knowledge sources, such as open-domain question answering, fact-checking, and slot filling.
Previous works mainly focus on probing PLMs whether they can memorize knowledge or perform well on downstream tasks. However, knowledge-related abilities are far beyond memorization. Thus, this work proposes a benchmark with diverse questions to evaluate memorization, identification, and reasoning ability. 
% In Table~\ref{benchmark_comparison}, we summarize the main characteristics of relevant benchmarks mentioned above.

\section{Knowledge Abilities Syllabus}
% \dou{using 1-2 sentences to introduce why we need such a syllabus}

% \dou{this section is very hard to read -- lots of new concepts are introduces. I suggest we change the writing style to: we propose evaluate XX from XX perspectives. ....}

To test the ability of PLMs, similar to the syllabus of SAT \footnote{https://leverageedu.com/blog/sat-syllabus/} designed for humans, in this section, we elaborate the design of the syllabus in terms of three perspectives, namely, 1) knowledge memorization; 2) knowledge identification, and 3) knowledge reasoning. These abilities correspond to \emph{System 1} and \emph{System 2} of the brain's reasoning system and are the basis of daily tasks~\cite{kahneman2011thinking}. In fact, \emph{System 1} covers the basic intuition and instinct, requiring excellent memorization and identification ability to accomplish simple routine works. \emph{System 2} is responsible for rational thinking and reasoning for complex decision-making activities.

\begin{table*}[t]
\centering
\caption{Knowledge Collection Schema. It displays the collected knowledge types, corresponding entity types and their examples.}
\label{schema}
\vspace{-0.3cm}
\resizebox{\textwidth}{!}{%
\begin{tabular}{clll}
\toprule
\multicolumn{2}{c}{Knowledge Type}                                                                  & \multicolumn{1}{c}{Entity Type}   & \multicolumn{1}{c}{Example}                 \\ \midrule
\multirow{7}{*}{\begin{tabular}[c]{c}Domain-Specific\\ Knowledge\end{tabular}}      & Physics   & physical quantity, state of matter   & The melting point of water is 0 $^o$C. \\
                                                                                & Biology   & biosystematics, biology habit        & A carnivore requires meat to survive.       \\
                                                                                & Chemistry & periodic table                       & Halogen is a Group 17 Element.              \\
                                                                                & Medicine  & pathology,drug effect                & The symptoms of frailty syndrome are osteoporosis.                                           \\ 
      & History   & cities in history, historical events & The capital of Serbian Despotate is located in the country Montenegro now. \\
                                                                                & Law       & articles of law                      & Treaty of Orihuela is a legislative act in the country Spain.  \\
                                                                                & Military  & weapon                               & M107 served during the war or conflict The Troubles.\\ \midrule
\multirow{2}{*}{\begin{tabular}[c]{@{}c@{}}General \\ Knowledge\end{tabular}} & Music     & musician, album, song                & Lady Gaga is a singer born on 1986 March 28.   \\
                                                                                & Film and TV      & actor, film, theme song              & The series The Wide Country's first broadcast was on 1962 September 20.  \\ \midrule
\multicolumn{2}{c}{Commonsense}                                                             & properties of common objects          & Rocks can be heavy.                 \\ \bottomrule
\end{tabular}%
}
\vspace{-0.3cm}
\end{table*}

\textbf{Knowledge Memorization.} Memorization characterizes the ability of remembering something about triples in the form of $<$ \texttt{subject}, \texttt{predicate}, \texttt{object} $>$ , where \texttt{subject} and \texttt{object} represent two entities, and \texttt{predicate} represents a relation between them. Similar to LAMA~\cite{Petroni2019LanguageMA}, we evaluate the models' memorization by triple completion task, that is, given two elements of a triple, the model is required to recover the remaining entity or relation. For example, given Joe Biden and the predicate \texttt{birthday}, it examines whether you remember the object \texttt{November 20, 1942}.

\textbf{Knowledge Identification.} Identification evaluates whether the PLMs can correctly recognize an entity or a triple. For entity-level identification, KMIR requires the model to distinguish the difference between entities in order to measure the representation ability of entities. For example, for several entities \texttt{Joe Biden, Barack Obama, Donald Trump, and Confucius}, in terms of the occupation, one might choose \texttt{Confucius} because the others are presidents of the U.S. To test such ability, we define a novel type of question called entity distinction. Given a set of entities, where one entity is different from the others, the model is supposed to pick the different one. For triple-level identification, the PLMs should tell whether a statement is true. This evaluation is rather important for current models, because many works~\cite{marcus2020next, lacker2020giving} have discovered the phenomenon of fabricating facts by the existing language models, e.g., GPT-3~\cite{brown2020language} tells that the sun has one eye. For this reason,  we design the questions called statement checking to validate the model's correctness toward a statement.

\textbf{Knowledge Reasoning.} KMIR measures the reasoning ability by checking whether the model can infer new knowledge from existing one. One of the most common knowledge reasoning paradigms for humans is predicate logical reasoning~\cite{atkinson1909art}. Thus, KMIR evaluates if the PLMs can perform basic predicate reasoning. Specifically, we refer to OWL2\footnote{https://www.w3.org/TR/owl2-syntax/} to require the model to infer new facts (if any) through classic object property axioms, including symmetry, transitivity, reflexive, equivalent, inverse, sub-relation, and SubOP(OPChain). Let us take an example for the axiom of transitivity, which is a general form of syllogism. Given two triples $<$\texttt{Michael Jackson}, \texttt{isA}, \texttt{singer}$>$ and $<$\texttt{singer}, \texttt{CapableOf}, \texttt{hold a concert}$>$, the model is supposed to infer $<$\texttt{Michael Jackson}, \texttt{CapableOf}, \texttt{hold a concert}$>$ (more details of axioms used in KMIR are listed in Sec.~\ref{q_generate}). To evaluate such inference ability, we automatically generate a lot of new knowledge equipped in the form of questions, then ask models to answer them.

\textbf{Knowledge Type.} So far, we have described the syllabus in terms of the abilities and related question types. Our syllabus also defines what types of knowledge are used for supporting the evaluation of such abilities. Here we follow the categorization of knowledge types defined in ~\cite{davis2015commonsense, storks2019recent, cambria2011isanette}, where the knowledge can be briefly classified into three types: general knowledge, commonsense, and domain-specific knowledge. The first two types which are often referred to in human communication, e.g., place of birth, country of citizenship, etc., are highly emphasized by previous benchmarks. They are suitable for measuring memorization and identification ability. 
In order to further promote the evaluation of reasoning ability, the domain-specific knowledge is particularly considered into our evaluation in the manner of  elementary-level natural science and social science knowledge to generate more diverse knowledge.

% \vspace{-0.3cm}
\section{Dataset Construction}
% \vspace{-0.3cm}
KMIR  is constructed by first collecting knowledge triples in diverse types and then converting them into cloze statements to test the language model. Specifically, the whole procedure contains four steps: 1) design knowledge collection schema, 2) collect knowledge triples, 3) generate questions, and 4) check the quality of questions. In the following, we illustrate each step in detail.

\subsection{Knowledge Collection Schema Design}
In the step of designing knowledge collection schema, we define the collected types of knowledge and types of entities. The idea of schema design is choosing a broad range of knowledge types to support the evaluation of abilities in our syllabus. Specifically, the knowledge in KMIR covers three categories as stated in previous section, i.e., general knowledge, domain-specific knowledge, and commonsense. \emph{General knowledge} collected in KMIR is mainly about entertainment information, including music and film \& TV knowledge. We collect the producer, release date, etc., of entities like famous songs, TV series or films, and basic information of entities like famous celebrities, etc. 
\emph{Domain-specific knowledge} in KMIR focuses on common elementary-level natural and social scientific knowledge, including physics, biology, chemistry, medicine, history, law, and military. We collect entities about basic scientific concepts for these domains, e.g., common physical quantities of substances, taxonomic knowledge of creatures, chemistry elements, medical drugs, cities in the history, articles of laws, information of weapons, etc. 
\emph{Commonsense} is mainly about properties of common objects, social conventions, etc. For instance, rocks can be heavy. Table~\ref{schema} displays examples of all knowledge types and related entity types in KMIR.

\begin{table*}[t]
\centering
% \vspace{-0.2cm}
\caption{Question templates for different knowledge types in KMIR.}
\vspace{-0.3cm}
\label{template}

\resizebox{\textwidth}{!}{%
\begin{tabular}{cllp{9.2cm}}
\toprule
\multicolumn{1}{c}{Knowledge Type}            & \multicolumn{1}{c}{Relation Type} & \multicolumn{1}{c}{Template}                                                                              & \multicolumn{1}{c}{Question}                       \\ \midrule
\multirow{5}{*}{\begin{tabular}{c}
   Domain-Specific  \\
   Knowledge
\end{tabular} }      &  comparison & Which of the following has \textless{}Comparison\textgreater \textless{}Physics Quantities\textgreater{}? <Choices> [MASK] & Which of the following has a higher melting point? ... \\
                                      &  capableOf          &     Which of the following can \textless{}capableOf\textgreater? <Choices>, [MASK].   &   Which of the following can be an element or an alloy? ...  \\
                                      &  melting point      &    The melting point of \textless{}chemical element\textgreater is [MASK] °C.  &    The melting point of water is [MASK] °C  \\ 
                                      &  \multirow{2}{*}{comparison}         &     \multirow{2}{*}{Which of the following has \textless{}Comparison\textgreater \textless{}Olympics games\textgreater{}? <Choices>, [MASK].}    &     Which of the following country held more times of the summer Olympics games? Canada, Switzerland, ..., [MASK].          \\
                                      &  capital of   &    \textless{}country\textgreater{}'s capital is  [MASK].      &    United States of America's capital is [MASK].            \\  \midrule
\multirow{2}{*}{\begin{tabular}{c}
   General \\
   Knowledge
\end{tabular}} &  grandfather of     &   Which of the following is the grandfather of \textless{}human\textgreater{}? <Choices>, [MASK].         &   Which of the following is the grandfather of Isaac Newton ...    \\
                                      &  spouseOf           &         \textless{}human\textgreater{} is spouse of [MASK].            &    Obama is spouse of [MASK].                   \\ \midrule
\multirow{1}{*}{Commonsense}          &  capableOf          &    [MASK] can \textless{}Ability\textgreater{}.       &  [MASK] can supply humans with milk.        \\ \bottomrule
\end{tabular}%
}
\vspace{-0.5cm}
\end{table*}

\subsection{Knowledge Collection}
\label{collection}
After obtaining the schema, we collect the three types of knowledge from two widely used knowledge bases, WikiData~\cite{Vrandecic2014WikidataAF} and ConceptNet~\cite{Speer2017ConceptNet5A}.

\textbf{WikiData.}
The general and domain-specific knowledge are collected from Wikidata. Specifically, we use the SPARQL query statement on the Wikidata Query Service platform \footnote{https://query.wikidata.org/} to retrieve the knowledge triples about interested entities.

\textbf{ConceptNet.}
The commonsense knowledge is collected from ConceptNet by using its official API \footnote{https://github.com/commonsense/conceptnet5/wiki/API} to extract the surface texts of interested entities, and then obtain the triples with relations depicting the properties of some common objects. We consider 7 relations: AtLocation, IsA, PartOf, HasA, HasProperty, UsedFor, and Causes.

In summary, by querying 95,532 interested entities with types defined in Table~\ref{schema} from the two knowledge bases, we obtain 192,078 knowledge triples covering diverse types of knowledge.

\subsection{Question Generation}
\label{q_generate}
KMIR contains 4 types of cloze questions, that is, triple completion, entity distinction, statement checking, and predicate reasoning, to evaluate the abilities in the proposed syllabus.

\textbf{Triple Completion Questions.}
To obtain questions evaluating knowledge memorization, KMIR uses a template-based method to convert the collected knowledge into cloze statements. We define a set of templates to convert triple into a natural language sentence based on its relation type and mask one entity to query the model. For example, the question template of birthday relation triple $<$\texttt{x}, \texttt{birthday}, \texttt{y}$>$ is "[x]'s birthday is [MASK]". Besides, motivated by ~\cite{jiang2020can}, each relation type corresponds to multiple versions of templates as prompts to make the questions more diverse. In Table~\ref{template}, we show some templates in KMIR. Note that, because subjects or objects in a triple could be multi-words, the models are supposed to tackle such cases that the [MASK] could be filled with multiple tokens. 

\textbf{Entity Distinction Questions.}
For entity distinction questions, KMIR provides natural language cloze-form prompts to require the model to pick one unique entity among the given four entities. For example, "James Cameron, Martin Scorsese, Peter Jackson, and Albert Einstein, which is an outlier in terms of occupation? [MASK]". In KMIR, the following types of entities are used for generating entity distinction questions: celebrities, famous scientists, chief and associate justices, chemical elements, species. We build hierarchical trees to category these entities based on their occupations, chemical properties, or biology taxonomy depicted in WikiData. Then, for generating such questions, we randomly sample three entities belonging to the same category and one entity from another. Finally, we design a set of templates to generate a cloze prompt based on these selected entities.

\textbf{Statement Checking Questions.}
For statement checking questions, we automatically generate corrupted knowledge triples to test if the models can distinguish them. Specifically, we replace the object or subject in a knowledge triple with other entities in the same relation to make up a corrupted one. For example, given a correct triple, $<$\texttt{water}, \texttt{melting point}, \texttt{0} $^o$C $>$, we replace the object \texttt{0 $^o$C} to \texttt{1538 $^o$C} which is the object in a similar triple  $<$\texttt{iron}, \texttt{melting point}, \texttt{1538 $^o$C}$>$. Remark that the corruption is further improved by filtering out the corrupted triples which have appeared in knowledge base.
Finally, in KMIR, 5,000 triples are selected from our collected triples to generate incorrect triples, and another 5,000 triples are selected as correct triples. Finally, we design some cloze-form prompts to query the model to check their correctness, e.g., "Is the statement `the melting point of water is 1538 $^o$C' true? [MASK]".

\textbf{Predicate Reasoning Questions.} 
For evaluating predicate logic reasoning, we firstly generate some new knowledge through rules based on OWL 2 Web Ontology Language Axioms~\cite{motik2009owl}. Then, we convert new knowledge to cloze questions. 6 types of axioms out of 14 are selected, and we design corresponding rules of knowledge generation for each axiom, as shown in Table~\ref{owl}. For example, to perform TransitiveOP, we sample commonsense knowledge about the occupation, e.g., (singer, CapableOf, hold a concert) from ConceptNet, and general knowledge from WikiData, e.g., (Michael Jackson, Is, singer), then infer the new knowledge (Michael Jackson, CapableOf, hold a concert).

\begin{table}
  \centering
%   \vspace{-0.4cm}
  \caption{Rules for generating new knowledge. $x$, $y$, $z$ are entity variables. $r, r_1, r_2$ indicate predicates.}
  \vspace{-0.4cm}
    \resizebox{7cm}{!}{    
    \begin{tabular}{cl}
    \toprule
    \multicolumn{1}{c}{Ontology Language Axiom} & \multicolumn{1}{c}{Rule Form} \\
    \midrule
    SymmetricOP & $(x, r, y) \rightarrow(y, r, x)$ \\
    TransitiveOP & $(x, r, y), (y, r, z) \rightarrow (x, r, z)$ \\
    % RelecxiveOP & $(x, r, x)$ \\
    EquivalentOP & $(x, r_1, y) \leftrightarrow (x, r_2, y)$ \\
    InverseOP & $(x, r_1, y) \rightarrow(y, r_2, x)$ \\
    Sub-relation OP & $(x, r_1, y) \rightarrow(x, r_2, y)$ \\
    SubOP(OPChain) & $(x, r_1, y),(y, r_2, z) \rightarrow(x, r, z)$ \\
    \bottomrule
    \end{tabular}%
    }
  \label{owl}%
  \vspace{-0.6cm}
\end{table}%

After obtaining the new knowledge, the questions can be generated in two ways. Some questions derived from new knowledge are obtained by simply masking one entity. However, in some cases, whether the subject or object in a triple is masked, the answer is not always unique. To avoid the multiple answers for such questions, we particularly design the cloze-form prompts by adding some incorrect answers.
For example, the question might be  "Which person is most likely to hold a concert, Michael Jackson, Albert Einstein, James Cameron, or Martin Scorsese?", where the last three entities are incorrect ones. In this case, the correct answer (i.e., Michael Jackson) is unique.
In total, we collect 184,348 questions.

% \vspace{-0.1cm}
\subsection{Quality Control}
% \vspace{-0.1cm}
\label{quality}
The automatic data generation method inevitably introduces some noises or controversial information into data, which could cause questions involving sensitive or unethical information and ambiguous answers. These noises are mainly derived from the core stages of question generation, including knowledge collection, template design, and automatic answer generation. 

Thus, we invite a group of annotators to manually check each sample's qualities in three main steps: 1) Filter out questions involving violence, pornography, discrimination, sovereignty disputes issues. 2) Correct the ambiguous questions by refining question templates. Sometimes, ambiguity is introduced by improperly querying the information. For example, for knowledge "Michael Jackson is born in USA.", if we query it by "Michael Jackson is born in [MASK].", the answer could be USA, Pennsylvania, etc. Thus, we refine the questions by introducing some phrases to restrict the answer scope, e.g., "\emph{The country} that Michael Jackson is born in is [MASK]". 3) Filter out triple completion questions whose answers are not unique. This is because in some cases, adding phrases still can not make the answer unique. For example, for the question "A [MASK] doesn't want a sore throat", the answer could be multiple, such as "human" or "singer", etc. This situation is very common in commonsense related questions. In summary, we randomly selected 16,000 questions to check and cost about 532 hours in total.

\subsection{Evaluation Metric}
As mentioned above, the answers in KMIR could be multiple tokens. KMIR hopes to measure how much the predictions coincide with the golden answers. Thus, following the commonly used metric~\cite{rajpurkar2016squad}, for each question, we split the prediction and correct answer into words separately, then calculate F1 measure between them.

\section{Experiments}

\begin{table*}[]
\centering
\caption{F1 scores (\%) of models on \textbf{different question types}. Note Random Guess performs 18.75\%  on predicate reasoning because these questions are a combination of multi-choice style clozes and open-end style clozes.}
\vspace{-0.4cm}
\label{exp_score}
\resizebox{15cm}{!}{%
\begin{tabular}{lcccccccc}
\toprule
\multicolumn{3}{c}{Method}               & & \multicolumn{4}{c}{Question Type}       \\ \cmidrule(r){1-3} \cmidrule(lr){5-8}
\multicolumn{1}{c}{Name}&\# Params & Training Corpus & & Triple Completion & Statement Checking & Entity Distinction& Predicate Reasoning                          \\ \midrule
Random Guess & -       &  -              & &    0.00017      & 50.00       & 25.00    & 18.75                                       \\ \midrule
BERT       & 110M      &   \multirow{4}{*}{BookCorpus \& Wikipedia} 
                                         & & 15.34        & 61.57       & 30.01    & 26.82                                      \\
ELECTRA    & 110M      &                 & & 14.98        & 63.09       & 33.27    & 27.52                                  \\
DistilBERT & 66M       &                 & & 14.37        & 59.89       & 28.56    & 24.38                                     \\ 
ALBERT     & 11M       &                 & & 11.79        & 63.42       & 32.46    & 27.16                                                        \\ \midrule
RoBERTa    & 125M      & \multirow{2}{*}{\begin{tabular}[c]{c}BookCorpus, Wikipedia, \\ OpenWebText, etc.\end{tabular}}  
                                         & & \textbf{15.50}        & \textbf{66.13}       & \textbf{35.92}    & \textbf{29.19}                                                    \\
DistilRoBERTa &  82M   &                 & & 14.64        & 63.02       & 32.67    & 27.02                                            \\ 
 \bottomrule
\end{tabular}%
}
% \vspace{-0.4cm}

\end{table*}

\subsection{Baselines}

We evaluate 3 groups of models, including simple baseline and state-of-the-art methods to investigate current PLMs' performances.

\begin{itemize}[leftmargin=*]
\vspace{-0.3cm}
\setlength{\itemsep}{2pt}
\setlength{\parsep}{-1pt}
\setlength{\parskip}{0pt}

\item Random Guess: This baseline aims to show the performance if we randomly pick a choice for open-ended stype cloze among all candidates answers, and for multi-choice style cloze among the choices shown in the questions.

\item BERT~\cite{devlin2019bert} \& its variants: BERT is a representative of a bidirectional encoder of Transformer. We also evaluate DistilBERT~\cite{sanh2019distilbert} to investigate the impact of PLM compression, ALBERT~\cite{lan2019albert} which is an enhanced version of BERT by introducing parameter-reduction techniques and a self-supervised loss, and ELECTRA~\cite{clark2020electra} which improves the BERT by introducing a more sample-efficient pre-training task called replaced token detection.

\item RoBERTa~\cite{liu2019roberta} \& its variants: RoBERTa improves BERT's training schema, e.g., larger batch size, sequence length. We also evaluate its compressed version, DistilRoBERTa~\cite{sanh2019distilbert}.
% \vspace{-0.3cm}
\end{itemize}

All the PLMs use the base version, and the train, dev, test set with 20,544, 144,085, 19,719 questions, respectively.

\subsection{Results}

\textbf{Overview.} Table~\ref{exp_score} summarizes the performances of the above baselines and display their number of parameters and training corpus. It can be seen that RoBERTa achieves the highest scores on all types of questions, but the results of all models are far from satisfactory (all below 70\%). It is still a formidable challenge for PLMs to achieve high memorization, identification, and reasoning abilities. In the following, we provide more analysis about the results.

\textbf{The Performances on Different Abilities.} 
Table~\ref{exp_score} shows the performances of baselines on different abilities. First, it should be noted that the values of the F1 score on identification (i.e., statement checking and entity distinction) and reasoning questions are relatively larger than memorization questions. It doesn't mean the reasoning is easier than memorization. This is because identification and reasoning questions contain some multi-choice-like cloze, while memorization questions have to provide an open-end answer. So, the answer spaces of different types of questions are different, and the scores between different abilities are incomparable.

Besides, from the ranking of models on triple completion scores, we can see that the largest PLMs (e.g., BERT, ELECTRA, and RoBERTa), achieve the highest scores, while the lightest weight PLM (i.e.,  ALBERT), obtains an obvious performance drop. It shows that the \textbf{models' performances on triple completion questions (i.e., memorization questions) are related to the number of parameters}. Besides, the enhanced versions of BERT (i.e., ELECTRA, ALBERT, RoBERTa) don't bring significantly improvements on memorization ability, but more obvious improvements on identification (i.e., statement checking and entity distinction) and reasoning ability. This shows that \textbf{improving the architecture or training schema of the BERT model mainly affects the knowledge identification and reasoning ability}.

\textbf{The Impact of Model Compression.} 
In Table~\ref{exp_score}, we compare the performances of models before and after model compression, e.g., BERT/RoBERTa vs. DistilBERT/DistilRoBERTa. We can see that model compression doesn't cause an obvious performance drop on memorization performance, but it hurts the performances more on the identification and reasoning questions. \textbf{This result may indicate that the parameters of PLMs could not only store the knowledge, but also store the relationship between knowledge.} 
It suggests that we may need other constraints or specific designs to maintain the reasoning ability when compressing the model parameters.

\begin{table}[h]
\centering
% \vspace{-0.1cm}
\caption{The prediction robustness F1 scores (\%) of models on Q-sets.}
% \vspace{-0.15cm}
\label{exp_consistancy}
\resizebox{7.5cm}{!}{%
\begin{tabular}{lccc}
\toprule
Method  & All Correct & Partially Correct & All Wrong \\ \midrule
BERT    & 16.37       & 54.28             & 29.35     \\
RoBERTa & 17.65       & 55.36             & 26.99     \\ \bottomrule
\end{tabular}%
}
% \vspace{-0.15cm}
\end{table}

\begin{table*}[ht]
\centering
\caption{F1 scores (\%) of models on questions involving \textbf{different types of knowledge}.}
\vspace{-0.4cm}
\label{exp_know}
\resizebox{\textwidth}{!}{%
\begin{tabular}{llccccccccccccc}
\toprule
\multirow{2}{*}{} & \multirow{2}{*}{Question Type} & \multicolumn{7}{c}{Domain-Specific Knowledge}      & \multicolumn{1}{l}{}  & \multicolumn{2}{c}{General Knowledge} & \multicolumn{1}{l}{} & \multicolumn{1}{l}{\multirow{2}{*}{Commonsense}} & \multicolumn{1}{l}{\multirow{2}{*}{Overall}} \\ \cmidrule(lr){3-9}   \cmidrule(lr){11-12}
                  &                                & Physics & Biology & Chemistry & Medicine &                       History     & Law    & Military    &                      & Music          & Film and TV          &  & \multicolumn{1}{l}{}                             & \multicolumn{1}{l}{}                         \\ \midrule
\multirow{3}{*}{\rotatebox{90}{{\small BERT}}}                  
                  & Triple Completion              &  5.95   & 19.28   & 12.10     & 11.30    &                       19.13       & 12.46  & 30.86       &                      & 13.59          & 10.31                &  &  5.71  & 15.34                                          \\
                  & Statement Checking                &  51.26  & 61.34   & 53.98     & 57.27    &                       64.57       & 66.32  & 71.23       &                      & 61.81          & 56.35                &  &  55.95 & 61.57                                            \\
                  & Reasoning                      &  10.74  &  11.42  & 27.78     & 14.45    &                       25.35       & 41.53  & 18.79       &                      & 41.38          & 50.64                &  & 24.56  &   26.82                                           \\ \midrule
\multirow{3}{*}{\rotatebox{90}{{\small RoBERTa}}}  
                  & Triple Completion              &  5.78   & 20.34   & 11.71     & 11.04    &                      19.51       & 11.67  & 31.03       &                      & 15.86          & 10.29                &  & 5.02   & 15.50                                              \\
                  & Statement Checking                &  50.42  & 70.00   & 55.65     & 56.36    &                      73.79       & 68.37  & 74.07       &                      & 57.80          & 61.11                &  & 58.33  & 66.13                                             \\
                  & Reasoning                      &  14.51  & 12.81   & 28.05     & 23.34    &                      27.92       & 41.99  & 23.49       &                      & 42.39          & 55.30                &  & 27.62  &  29.19                                            \\ \bottomrule
\end{tabular}%
}
% \vspace{-0.4cm}
\end{table*}

\textbf{Robustness of PLM Prediction.}
In KMIR, given one knowledge triple, we construct multiple types of cloze questions to investigate different knowledge-related abilities. For example, for a triple <Joe Biden, birthday, November 20, 1942>, the questions might be "Joe Biden's birthday is [mask]" and "Joe Biden's birthday is November 20, 1942, is that right?"
A robust model needs to be able to maintain the high consistency of the answers to these highly relevant questions. Thus, we group the questions derived from the same knowledge triple together as one question set (denoted as Q-set). Then, we calculate how many q-sets are partially correctly as shown in Table \ref{exp_consistancy}. 
It can be seen that more than 50\% of Q-sets for BERT and RoBERTa are partially correctly answered. \textbf{This result reflects that current models are hard to robustly store the knowledge and validate whether the saved knowledge is correct}.

\begin{figure}[t]
    \centering
    \includegraphics[width=7.0cm]{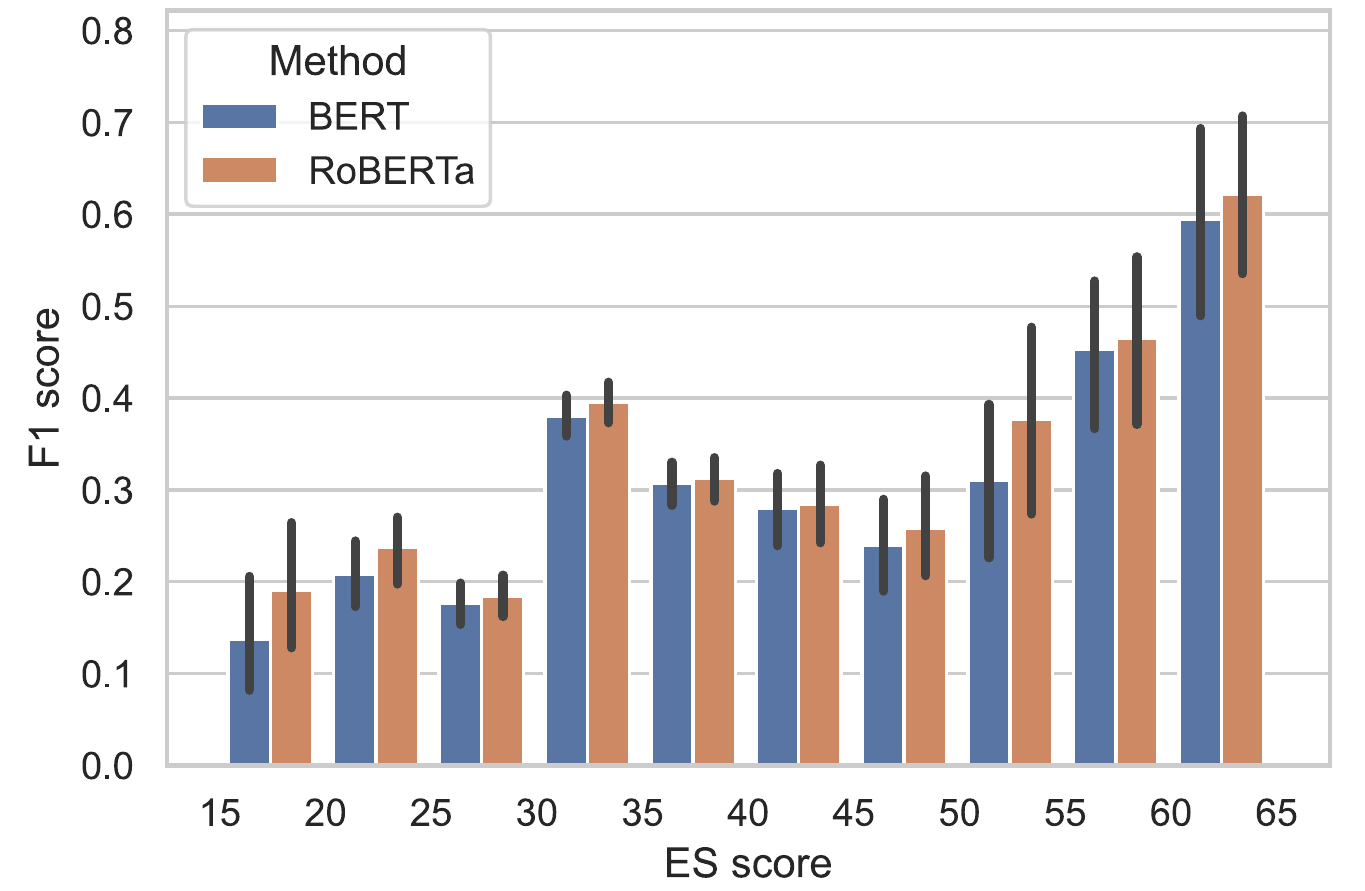}
    % \vspace{-0.4cm}
    \caption{The correlations between F1 scores of BERT/RoBERTa and different ES scores on questions generated by the new knowledge under the rules TransitiveOP and SubOP(OPChain).}
    \label{es_score}
    % \vspace{-0.5cm}
\end{figure}

\textbf{Can the PLM really Reason?}
KMIR uses a rule-based method to generate a large amount of knowledge that is not explicitly expressed in the training corpus. From results in Table~\ref{exp_score}, PLMs perform relatively well on these reasoning questions. Is PLM really able to infer new knowledge, or does it simply memorize the co-occurrence between words? For example, although no Wikipedia page directly tells Michael Jackson can record a song, the words \emph{Michael Jackson} and \emph{song} might appear within sentences or paragraphs many times. Thus, we need investigate whether PLMs mainly learn co-occurrence between tokens rather than new knowledge. 

We establish an Elastic Search system~\cite{divya2013elasticsearch} by indexing all Wikipedia pages , and then compute the co-occurrence score (ES score) between entities in a query and the words in the corresponding answer appearing in Wikipedia pages. We display the correlation between 
performances of models on questions where the new knowledge is generated by rules TransitiveOP and SubOP(OPChain), and the ES scores in Figure~\ref{es_score}. It can be seen that, in general, current models perform better on questions with higher ES scores. \textbf{This suggests that models 
 who can answer reasoning questions in that they just memorized context information by a large-amount of parameters.} However, the co-occurrence cannot solve all reasoning paradigms. We believe that there is still much room for explicit reasoning for existing PLMs.

\textbf{The Performances on Different Knowledge Types.} 
Finally, it remains to check the performances of all models with respect to different types of knowledge. From Table~\ref{exp_know}, we can see that the performances vary largely in different types of knowledge. Note that, the entity distinction questions are not listed because the entities in one question involve multiple types of knowledge. The results indicate that performances over knowledge-related abilities are different among knowledge types. The memorization of commonsense and physics is relatively harder. 
Reasoning questions about domain-specific knowledge are more difficult than general knowledge. \textbf{These results also show that current models are hard to precisely memorize the facts and implement complex reasoning.}

\vspace{0.1cm}
\section{Conclusion}
\label{conclusion}
\vspace{0.1cm}
This paper introduces KMIR benchmark for evaluating knowledge memorization, identification, and reasoning abilities. KMIR includes a systematic evaluation syllabus to summarize the knowledge-related abilities of PLMs, and has 184,348 questions involving 4 types of questions covering 3 types of knowledge. We also find many interesting phenomena through extensive experiments: 1) The memorization ability of PLMs depends more on the number of parameters than training schemes. 2) Current PLMs are struggling to robustly remember the facts. 3) Model compression technology retains the amount of knowledge well, but hurts the identification and reasoning ability, etc.

\bibliography{anthology}
\bibliographystyle{acl_natbib}

\clearpage

\begin{onecolumn}

\appendix

\section*{Appendix}
\label{sec:appendix}

% This is an appendix.
In the appendix, we provide more details about dataset construction (Sec.~\ref{supp_cons}), data statistics (Sec.~\ref{supp_stat}), implementation of baselines (Sec.~\ref{supp_detail}), and license and usage of KMIR (Sec.~\ref{supp_license}).

\section{Dataset Construction}
\label{supp_cons}

In this section, we provide more examples of the selected knowledge triples and question templates used for generating  questions.  

\subsection{Knowledge Collection Schema}
In the Table~\ref{supp_schema}, we show more knowledge collection schema.

\begin{table*}[h]
\centering
\caption{Knowledge Collection Schema. This table displays our collected 10 specific types of knowledge and corresponding relation types and their examples.}
\label{supp_schema}
\resizebox{\textwidth}{!}{%
    \begin{tabular}{cccp{10cm}}
    \toprule
    \multicolumn{2}{c}{Knowledge Type} &       Relation Type & \multicolumn{1}{c}{Example} \\
    \midrule
    \multirow{15}[2]{*}{\begin{tabular}{c}
        Domain-specific \\
        Knowledge
    \end{tabular} } 
          & \multirow{4}{*}{Physics} & physical quantity & The melting point of water is 0 centigrade. \\
          &       & physical property & A metal can conduct electricity and heat well. \\
          &       & \multirow{2}{*}{state of matter} & Rocks has a stable, definite shape and volume at standard temperature and pressure. \\ \cmidrule{2-4}
          & \multirow{2}{*}{Biology} & biosystematics & Homeothermic species maintain a stable body temperature. \\
          &       & biology habit & A carnivore requires meat to survive. \\ \cmidrule{2-4}
          & \multirow{2}{*}{Chemistry} & periodic table & Halogen is a Group 17 Element. \\ 
          &       & atomic number & aluminium's atomic number is 13. \\ \cmidrule{2-4}
          & \multirow{2}{*}{Medicine} & pathology & The symptoms of frailty syndrome are osteoporosis. \\
          &       & cause & HIV infection is caused by the human immunodeficiency virus. \\ \cmidrule{2-4}
    
          & \multirow{3}{*}{History} & \multirow{2}{*}{cities in history}  & The capital of Serbian Despotate is located in the country Montenegro now. \\
          &       & historical events & The American Civil War was a civil war in the United States. \\ \cmidrule{2-4}
          & \multirow{2}{*}{Law}   & articles of law & Treaty of Orihuela is a legislative act in the country Spain. \\
          &       & lawer & Willem Eduard Bok Jr. was a South African lawyer. \\ \cmidrule{2-4}
          & \multirow{3}{*}{Military} & weapon & M107 served during the war or conflict The Troubles. \\
          &       & \multirow{2}{*}{manufacturer} & Northrop Grumman Corporation is an American multinational aerospace and defense technology company.  \\
    \midrule
    \multirow{6}[2]{*}{General
Knowledge} & \multirow{2}{*}{Music} & musician & Lady Gaga is a singer born on 1986 March 28. \\
          &       &  album, song & One Love is a song by Justin Bieber released in album Believe. \\  \cmidrule{2-4}
          & \multirow{4}{*}{Film and TV} & \multirow{2}{*}{actor, film} & The series The Wide Country’s first broadcast was on 1962 September 20. \\ 
          &       & \multirow{2}{*}{theme song} & The theme musics of the television series Friends is I'll Be There for You. \\
    \midrule
    \multicolumn{2}{c}{\multirow{4}[2]{*}{Commonsense}} & CapableOf & Rocks can be heavy.  \\ 
    \multicolumn{2}{c}{} & IsA   & eight ball is a game of geometry. \\
    \multicolumn{2}{c}{} & HasProperty & Africa can be one of the largest continents. \\
    \multicolumn{2}{c}{} & HasA  & The snake have a nest full of babies,scales,no legs. \\
    \bottomrule
    \end{tabular}%
}
\end{table*}

\subsection{Question Generation}
In Table~\ref{tab_q_type_sc_ed}, we display the question templates used for generating the entity distinction and statement checking questions.

\begin{table}[h]
\centering
\caption{Question templates for entity distinction and statement checking questions.}
\label{tab_q_type_sc_ed}
\resizebox{\textwidth}{!}{%
\begin{tabular}{cp{16cm}}
\toprule
Question Type      & \multicolumn{1}{c}{Question Template}                                                                                                                                                                                                                               \\ \midrule
\multirow{4}{*}{Entity Distinction} & Is the statement “\textless{}Statement of a knowledge triple\textgreater{}” true? {[}MASK{]}.                                                                                                                                                                         \\
                   & \textless{}Statement of a knowledge triple\textgreater{}, is this true? {[}MASK{]}.                                                                                                                                                                              \\
                   & Is this true, \textless{}Statement of a knowledge triple\textgreater{}? {[}MASK{]}.                                                                                                                                                                                  \\
                   & Is the following statement true, \textless{}Statement of a knowledge triple\textgreater{}? {[}MASK{]}.                                                                                                                                                               \\ \midrule
\multirow{6}{*}{Statement Checking} & \textless{}Entity1\textgreater{}, \textless{}Entity2\textgreater{}, \textless{}Entity3\textgreater{}, and \textless{}Entity4\textgreater{}, which is a outlier, in terms of the \textless{}Aspect\textgreater{}? {[}MASK{]}.                                         \\
                   & Which is different with others in terms of the \textless{}Aspect\textgreater{}, \textless{}Entity1\textgreater{}, \textless{}Entity2\textgreater{}, \textless{}Entity3\textgreater{}, and \textless{}Entity4\textgreater{}? {[}MASK{]}.                              \\
                   & Which is different with others among the following entities in terms of the \textless{}Aspect\textgreater{}, \textless{}Entity1\textgreater{}, \textless{}Entity2\textgreater{}, \textless{}Entity3\textgreater{}, and \textless{}Entity4\textgreater{}? {[}MASK{]}. \\
                   & Among \textless{}Entity1\textgreater{}, \textless{}Entity2\textgreater{}, \textless{}Entity3\textgreater{}, and \textless{}Entity4\textgreater{}, {[}MASK{]} is the outlier in terms of the \textless{}Aspect\textgreater{}.                                        \\ \bottomrule
\end{tabular}%
}
\end{table}

For reasoning questions, we first generate lots of new knowledge according OWL 2 Web Ontology Language Axioms~\cite{motik2009owl}, then diverse question templates are designed for each relation type of the knowledge fact. In Table~\ref{supp_q_template}, we display the examples of generated new knowledge for each axiom in OWL 2. Table~\ref{supp_template} shows our collected types of relations and corresponding templates for triple completion and predicate reasoning questions.
% Table generated by Excel2LaTeX from sheet 'OWL'
\begin{table}[htbp]
  \centering
  \caption{Examples of generated new knowledge for each axiom in OWL 2. $x$, $y$, $z$ are entity variables. $r$ indicates a predicate.}
  \label{supp_q_template}%
  \resizebox{\textwidth}{!}{

    \begin{tabular}{llp{11cm}}
    \toprule
    \multicolumn{1}{c}{Axiom} & \multicolumn{1}{c}{Rule Form} & \multicolumn{1}{c}{Example} \\
    \midrule
    SymmetricOP & $(x, r, y) \rightarrow(y, r, x)$ & (peace, oppsite of, war) $\rightarrow$ (war, oppsite of, peace) \\
    TransitiveOP & $(x, r, y), (y, r, z) \rightarrow (x, r, z)$ & (cat, isA, mammal),(mammal, isA, vertebrate) $\rightarrow$ (cat, isA, vertebrate) \\
    EquivalentOP & $(x, r_1, y) {\leftrightarrow} (x, r_2, y)$ & (Joe Biden, is the citizen of, USA) $\leftrightarrow$ (Joe Biden, nationality, USA) \\
    InverseOP &  $(x, r_1, y) \rightarrow(y, r_2, x)$ & (USA, capital, Washington) $\rightarrow$ (Washington, country, USA) \\
    Sub-relation OP & $(x, r_1, y) \rightarrow(x, r_2, y)$ & (Gordian I, father of, Gordian II)$\rightarrow$ (Gordian I, parent of, Gordian II) \\
    \multirow{2}{*}{SubOP(OPChain)} & \multirow{2}{*}{$(x, r_1, y),(y, r_2, z) \rightarrow(x, r, z)$} & (Napoleon, country, French),(French, continent, Europe) $\rightarrow$ (Napoleon, continent, Europe) \\
    \bottomrule
    \end{tabular}%
    }
\end{table}%

\begin{table}[hb]
\centering
\caption{Question templates for different types of knowledge in KMIR.}
\label{supp_template}
\resizebox{\textwidth}{!}{%
    \begin{tabular}{ccp{10cm}p{10cm}}
    \toprule
    Knowledge Type & Relation Type & \multicolumn{1}{c}{Template} & \multicolumn{1}{c}{Question} \\
    \midrule
    \multirow{14}[2]{*}{\begin{tabular}{c}
        Domain-specific \\
        Knowledge
    \end{tabular}} & comparison & Which of the following has <Comparison><Physics Quantities>? <Choices>, [MASK]. & Which of the following has a higher boiling point? methanol,carbon tetrachloride,methyl formate,water,  \\
          & capableOf & Which of the following can <capableOf>? <Choices>, [MASK]. & Which of the following can be an element or an alloy? \\
          & melting point  & The melting point of <chemical element>is [MASK] °C. & The melting point of water is [MASK] °C \\
          & IsA   & Which of the following <IsA>is a <Physics Quantities>? <Choices>, [MASK]. & which of the following is a Group 17 Element? \\
          & HasProperty & Which of the following has <HasProperty><Physics Property>? <Choices>, [MASK]. & Which material in the following has a stable, definite shape and volume at standard temperature and pressure? \\
 & comparison & Which of the following has <Comparison><Olympics games> <Choices>, [MASK]. & Which of the following country held more times of the summer Olympics games?  \\
          & capital of & <country>’s capital is [MASK].  & United States of America’s capital is [MASK]. \\
          & inception & The historical country <historical country> was established on [MASK]. & The Mongol Empire was established in the 1206. \\
          & countryOf & <historical period> was a period  in the country [MASK]. & Norman architecture was a historical period of the country Fracne \\
    \midrule
    \multirow{3}[2]{*}{General
Knowledge} & grandfather of & which of the following is the grandfather of <human>? & which of the following is the grandfather of Isaac Newton ? \\
          & spouseOf & <human> is spouse of [MASK]. & Obama is spouse of [mask]. \\
          & FatherOf & [MASK] is <human>'s father. & Gordian I is Gordian II's father. \\
    \midrule
    \multirow{2}[2]{*}{Commonsense} & capableOf & <entity> can <capbleOf>. & a fact can be [MASK]. \\
          & OppsiteOf & Which of the following is the oppsite of <entity> ? & ocean is oppsite of continent. \\
    \bottomrule
    \end{tabular}%

}
\end{table}

\section{Data Statistics}
\label{supp_stat}

Figure~\ref{supp_statistic} visualizes the question distribution for different domains and abilities. The results show that KMIR has a relatively uniform distribution regardless of question type or knowledge type.
\begin{figure}[ht]
    \centering
    \includegraphics[width=\textwidth]{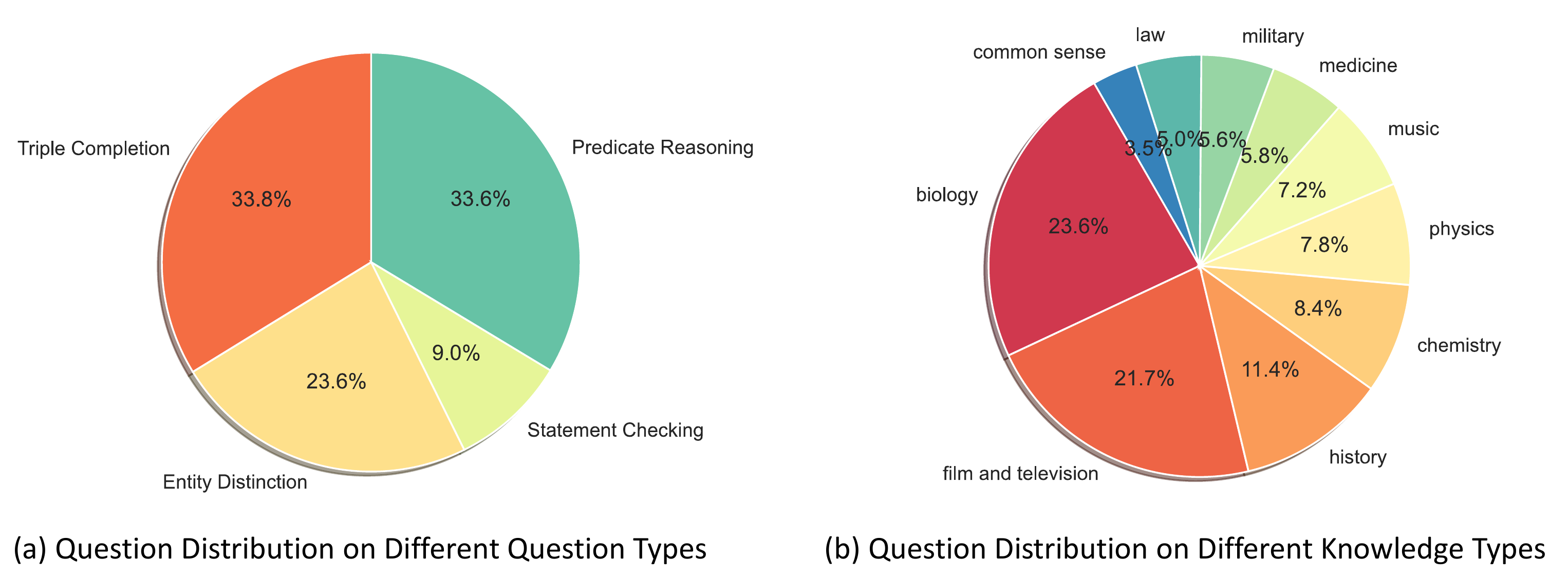}
    \caption{Statistics of questions in KMIR.}
    \label{supp_statistic}
\end{figure}

\section{Implementation Details of Baselines}
\label{supp_detail}

KMIR requires the PLMs to output multiple tokens as the answer. To adjust current PLMs to fit this setting, we modify the inputs of PLMs in fine-tuning. Specifically, we expand the [MASK] in a query into 8 masks (the maximum number of masked tokens in KMIR). For example, the original query "Singer Don Omar 's birthday is [MASK]" is transformed into "Singer Don Omar 's birthday is [MASK] [MASK] [MASK] ...". Similarly, the correct answer is transformed into tokens, then padded into 8 tokens for training the models. Besides, to better reflect the knowledge-related abilities, the PLMs are fine-tuned on each type of questions and test on corresponding test set questions.

\section{License and Usage}
\label{supp_license}

Firstly, in KMIR, the personnel information involved in the dataset is about public information of public figures from WikiData and does not involve personal privacy. Then, the questions in KMIR are under CC BY-NC-SA 4.0 license. For the knowledge base that we used to generate questions, WikiData is licensed under the Creative Commons CC0 License, and ConceptNet is under a Creative Commons Attribution-Share Alike 4.0 International License. Besides, our dataset collects the knowledge from open and free knowledge base, the knowledge used in dataset could have some mistakes or out-of-date. Finally, our data can only be used for academic researches and do not support commercial usages.

\end{onecolumn}

\end{document}